\documentclass[runningheads]{llncs}
\usepackage{graphicx}

\usepackage{tikz}
\usepackage{comment}
\usepackage{amsmath,amssymb} 
\usepackage{color}

\usepackage[accsupp]{axessibility}  

\usepackage{bbm}
\usepackage{bbding}
\usepackage{multirow}
\usepackage{booktabs}
\newcommand{\norm}[1]{\left\lVert#1\right\rVert}
\usepackage{cite}

\begin{document}
\pagestyle{headings}
\mainmatter
\def\ECCVSubNumber{3520}  

\title{Interclass Prototype Relation\\ for Few-Shot Segmentation} 

\titlerunning{Interclass Prototype Relation for Few-Shot Segmentation}
%

\author{Atsuro Okazawa}
\authorrunning{A. Okazawa}
%

\institute{R\&D Promotion Office AI Strategy Office,\\ SoftBank Corp., Tokyo, Japan\\
\email{atsuro.okazawa@g.softbank.co.jp}}
\maketitle

\begin{abstract}
Traditional semantic segmentation requires a large labeled image dataset and can only be predicted within predefined classes.
To solve this problem, few-shot segmentation, which requires only a handful of annotations for the new target class, is important.
However, with few-shot segmentation, the target class data distribution in the feature space is sparse and has low coverage because of the slight variations in the sample data.
Setting the classification boundary that properly separates the target class from other classes is an impossible task. In particular, it is difficult to classify classes that are similar to the target class near the boundary.
This study proposes the Interclass Prototype Relation Network (IPRNet), which improves the separation performance by reducing the similarity between other classes.
We conducted extensive experiments with Pascal-$5^i$ and COCO-$20^i$ and showed that IPRNet provides the best segmentation performance compared with previous research.
\end{abstract}
\section{Introduction}
\begin{figure}[h]
  \includegraphics[width=1.0\textwidth]{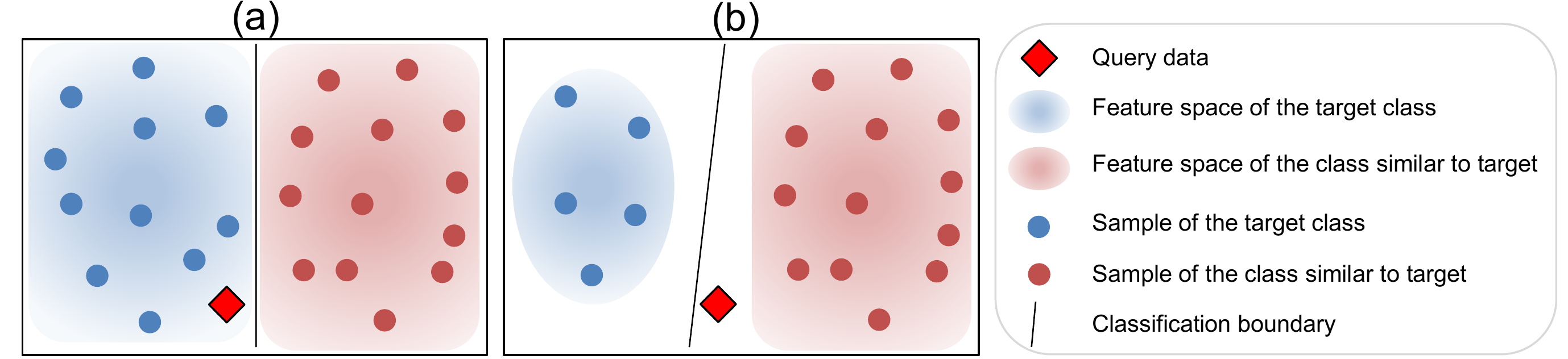}
  \caption{Each shade is the area to which most of the samples in that class are mapped. The black line is the classification boundary calculated from the samples.
  (a) shows the area to be mapped from sufficient samples is given, not in the condition of few-shot. (b) shows the area mapped in the few-shot problem with a few samples.
  If query data is plotted near the boundary between the two classes you want to classify, we cannot detect the class that originally belonged to the target class in the case of few-shot with a narrow-mapped area.
  \label{fig:fg1}
}
\end{figure}

Recent advances in semantic segmentation have been brought about by advanced convolutional neural networks (CNNs)~\cite{DNN} and large labeled image datasets~\cite{pascal}, ~\cite{coco},~\cite{city},~\cite{ade}.
However, semantic segmentation with fully supervised learning requires a substantial number of annotations per pixel and can be time-consuming to create.
To solve this problem, few-shot segmentation that requires only a handful of annotations for a new target class is important.
Few-shot segmentation aims to obtain generalization ability from known classes and adapt them to new target classes via a few shots, namely, support data.
However, few-shot segmentation is not under the condition in which features can be extracted from a large amount of data with all variations (Fig.\ref{fig:fg1}(a)), and the target class data distribution in the feature space is sparse and has low coverage (Fig.\ref{fig:fg1}(b)).
Therefore, there is an essential problem in that it is not possible to set the classification boundary that separates the target class from other classes properly. In particular, it is difficult to classify classes that have features like those of the target class near the boundary.
To tackle this important problem without increasing the number of shots for the target class, it is important to differentiate the features between each class 
when learning generalization abilities from known classes.

Few-shot segmentation is an extension of the technology based on few-shot learning~\cite{MatchNet},~\cite{RelationNet},~\cite{Prototype},~\cite{Learning2Learn},~\cite{MAML},~\cite{Optimization}, and it tackles the more difficult task of predicting the label for each pixel, instead of predicting a single label for the entire image in few-shot learning. Few-shot learning has meta-learning~\cite{Learning2Learn},~\cite{MAML} and metric learning~\cite{MatchNet},~\cite{RelationNet},~\cite{Prototype} as the mainstream methods. Meta-learning was introduced by Shaban et al.~\cite{Oneshot} and metric learning by Snell et al.~\cite{Proto_seg} for segmentation problems. In particular, the metric learning approach to the problem of few-shot segmentation has been actively studied in recent years and has been successful. Our research is related to few-shot segmentation using metric learning.
The method of few-shot segmentation using metric learning has been pushed into a global descriptor called a prototype using supporting data~\cite{AMP}. The support data are a few samples with a target class. The prototype is a vector representation of the features for the target class and there are many studies based on the method of inference by comparing the prototype with the features of the query image.
For better prototyping and proper comparison of support and query, there are several earlier studies. For example, there are studies that have introduced a mechanism to separate the foreground and background~\cite{PANet}, ~\cite{ASR}, ~\cite{MLC}, and studies that have introduced a multi-scale architecture~\cite{PGNet},~\cite{PFENet},~\cite{ASGNet}.
In these studies, the prototype extracted from the support data was appropriately compared with the query data, and training was performed based on the loss function between the query data and its ground truth.

However, in few-shot segmentation with only a few shot samples, as mentioned above, there are few variations of support data using the target class, and prototype generation is performed from a sparse feature space.
Therefore, it is particularly difficult to obtain a prototype that can classify classes with features similar to those of the target class. However, owing to the problem setting of the few-shot, it is not possible to increase the amount of data for the target class and make the feature space dense.

Accordingly, we propose the Interclass Prototype Relation Network (IPRNet) that improves separation performance by reducing the similarity between types to highlight the differences between the prototypes of similar classes. IPRNet has 1) the Interclass Prototype Relation Module, which aims to improve the separation performance between similar classes by reducing the similarity between prototypes of each class, and 2) the Respective Classifier Module, which aims to improve the separation performance by integrating respective estimations of the target class and background. We hypothesized that these modules could improve the separation performance of the target and other classes. In this study, we verified this hypothesis using two experiments. First, we evaluated whether the performance would improve compared to earlier research with the best performance. Second, we conducted an ablation study to verify that the modules  proposed in IPRNet are effective in classifying similar classes. The contributions of this study are as follows.
\begin{itemize}
\item We propose a novel few-shot segmentation method called IPRNet, which improves the separation performance between the target class and other classes that are especially similar to the target.
\item We evaluate Pascal-$5^i$\cite{Oneshot} and COCO-$20^i$~\cite{coco} and show that the proposed method improves the mean intersection over union (mIoU) over the existing best-performing method.
\item Through an ablation study, we verified whether the proposed method is effective for classifying similar classes.
\end{itemize}
Our code is at:\\\url{http://fss-stat.s3-website-us-east-1.amazonaws.com/IPRNet.zip}
\section{Related Work}
Few-shot segmentation mostly consists of few-shot learning-based technology that improves model generalization ability and semantic segmentation technology that solves pixel-level classification problems.
We describe the existing research related to these constituent requirements and issues.
\subsubsection{Semantic Segmentation}
In semantic segmentation, deep neural networks based on convolutional neural networks (CNNs)~\cite{DNN} have been successful.
Starting from fully convolutional networks~\cite{FCN}, especially  encoder-decoder structure proposed by Segnet~\cite{Segnet} has become the basic network structure in recent semantic segmentation.
Recently, a faster method Enet~\cite{Enet}, and encoder-decoder structures that ensemble multi-scale features to express all frequency information have been proposed~\cite{DilatedConv},~\cite{Deeplab},~\cite{Rethink_deeplab},~\cite{Deeplabv3+},~\cite{PSPNet},~\cite{ICNet}.
The latest research also proposed a convolution-free and resolution deterioration-free method~\cite{TformerSS} based on the transformer~\cite{Transformer}, which is a model that uses only the attention mechanism instead of CNNs~\cite{DNN}.
\subsubsection{Few-shot Learning}
Few-shot learning focuses on the generalization ability of the model and enables learning for new class predictions using a few annotated samples.
The mainstream existing methods are metric learning\cite{MatchNet},~\cite{RelationNet},~\cite{Prototype} and meta-learning~\cite{Learning2Learn},~\cite{MAML},~\cite{Optimization}. 
The core idea of metric learning is the distance measurement, which is formulated as an optimization of the distance or similarity between the images and regions. Meta-learning focuses on achieving a high-speed learning ability by defining specific optimization functions and loss functions. 
Among these methods, the concept of a prototypical network~\cite{Prototype} is widely adopted for few-shot segmentation, and it is possible to reduce the calculation cost significantly while maintaining high performance.
Many methods focus on image classification, but recently few-shot segmentation has attracted attention.
\subsubsection{Few-shot Segmentation}
Few-shot segmentation is an extension of few-shot learning that addresses the more difficult task of predicting a label for each pixel rather than predicting a single label for the entire image.
Few-shot meta-learning was introduced into the segmentation problem by Shaban et al.~\cite{Oneshot}, and there is a lot of research on its enlargement~\cite{Differentiable},~\cite{Init},~\cite{Without},~\cite{CWT}. 
Few-shot metric learning was successfully introduced by Snell et al.~\cite{Proto_seg}.
Many of the methods so far drop the problem into a 1-way classification problem in order to apply it to episodic learning~\cite{MatchNet} to acquire generalization ability. 
In previous research, they pushed the support data into a global descriptor to obtain a prototype that is the features of the class in the first step~\cite{AMP}. 
In the next step, the target object and background are separated by comparing the prototype with the query image~\cite{FWB},~\cite{CANet},~\cite{PANet},~\cite{PGNet},\cite{PPNet},~\cite{PMMs},~\cite{PFENet},~\cite{ASGNet},~\cite{MLC}.
In addition to these, research to absorb the difference in size, position, and orientation of the target object on the support image and the query image, and to compare them correctly have been conducted~\cite{CRNet},~\cite{FGN},~\cite{SCL},~\cite{DAN},~\cite{SAGNN}.
There is also a method to infer from the correlation between all positions of query data and support data~\cite{HSNet}.
However, most of them are solved by general $1$-way classification problems, so only the relationship between the target class and the background can be considered. ASR~\cite{ASR} is a method that uses multiple latent class vectors, but the feature map channel is divided and assigned to each class.
Therefore, if many classes are included, the number of channels assigned to one class will decrease and it will not work effectively.
Existing methods do not fully consider the relationships between different classes, making proper classification difficult.
In particular, it is the most difficult to separate from similar classes near the discriminant boundary, and  to the best of our knowledge, no research has been conducted on this problem.
This research proposes a novel IPRNet that focus on improvement between similar classes, which are particularly difficult to classify.
\section{Problem Definition}
The major difference between few-shot segmentation and general semantic segmentation is that the training and test set categories do not intersect.
Particularly, at the inference stage, the test set had classes that were not found during the training.
Specifically, given the train set $S_{train} = {\{(I^{S/Q},M^{S/Q})\}}$ and the test set $S_{test} = {\{(I^{S/Q}, M^{S/Q})\}}$, the categories of the two sets do not intersect $(S_{train}\cap S_{test}=\phi)$.
Here, $I\in R_{H\times W\times 3}$ shows an RGB image and $M\in R_{H\times W}$ shows a segmentation mask.
The subscripts $S$ and $Q$ represent support and query, respectively.
We mimicked the first one-shot segmentation study~\cite{Oneshot} and applied training and testing to an episodic learning framework.
In each episode, the input to the model consists of the query image $I^Q$ and k samples $(I^{S}_i , M^{S}_i)$,
$i\in\{ 1, 2, \ldots, k \}$ from the support set.
All support image and query image have the same class $c$.
Training selects $(I^Q, M^Q, I^{S}_i , M^{S}_i)$ of 
batch size $b$ set from the train set $S_{train} = {\{(I^{S/Q},M^{S/Q})\}}$ and estimates the query mask $\tilde{M^Q}$ to approximate ground truth mask $M^Q$.
\section{Proposed Method}
\subsection{Design guideline of network structure}
\begin{figure}[t]
  \includegraphics[width=1.0\textwidth]{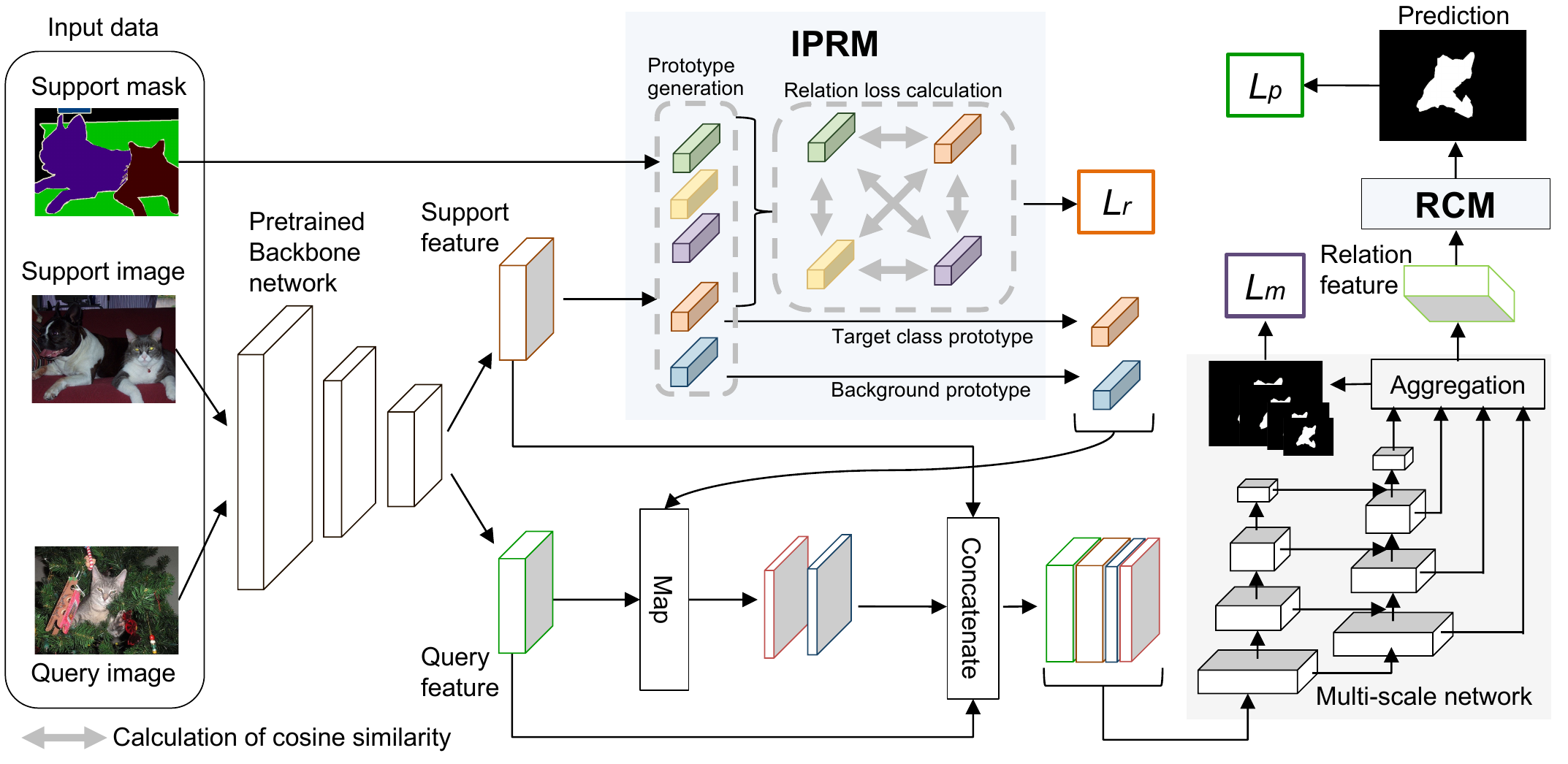}
  \caption{The Overall architecture of the proposed method, Interclass Prototype Relation Network (IPRNet).}
  \label{fig:overrall}
\end{figure}

As mentioned in the introduction, the target class, which has a few
shots in few-shot segmentation, is difficult to classify similar classes because the data plotted in the feature space is sparse and has low coverage.
To address this problem, we propose the Interclass Prototype Relation Network (IPRNet).
IPRNet has two modules: the Interclass Prototype Relation Module (IPRM) and the Respective Classifier Module (RCM).
These modules aim to improve the identification performance of similar classes by reducing the similarity between prototypes and extracting the differences between classes.
An overview of the network is shown in Figure \ref{fig:overrall}.
\subsection{Interclass Prototype Relation Network}
This section describes the overall flow of IPRNet.
First, the support and query images were fed into pretrained shared CNNs (pretrained by ImageNet~\cite{ImageNet}) to extract features.
Next, we passed the support features with the support masks through the IPRM. Through IRPM, we obtain prototypes that represent feature vectors for each class, and the value $L_r$ that indicates the similarity between each prototype.
Then, for more accurate pixel-by-pixel matching, the matching between prototypes and the query feature is performed by calculating the cosine similarity in map process.
The two similarity maps can be obtained by matching with the query feature for the target class prototype and the background prototype, respectively.
This process was inspired by earlier research on MLC~\cite{MLC}.
The input to the multi-scale network is the concatenation of the support features, the query feature, and two similarity maps. The output is a relation feature valid for classification and $L_m$ which is a multi-scale loss introduced in PFENet~\cite{PFENet}.
The multi-scale network sets up the top-down structure of FPN-like~\cite{FPN} by using the feature enrichment module introduced in PFENet~\cite{PFENet} to obtain multi-scale information.
This structure enables fast multi-scale aggregation, by transferring features from finer to coarser and by easing feature interaction.
Each scale yielded segmentation results for calculating the loss. The average loss value for each scale was $L_m$.
Finally, the relation feature, a multi-scale information-intensive feature map, is generated by fusing all the different scales into a concatenated feature map by convolution.

The relation feature is the input to the RCM.
The RCM performs a discriminative process to classify the foreground and background more and obtains the final inference result $\tilde{M^Q}$. Then, the loss $L_p$ between the inference result and ground-truth mask $M^Q$ is calculated.
The loss function, which is the cost function of training, is given by equation (\ref{eq:loss_f}).
\begin{align}
\label{eq:loss_f}
Loss = w_1 L_r + w_2 L_m + w_3 L_p
\end{align}
$w_1$, $w_2$, and $w_3$ are the weight coefficients, which were trained with $w_1=0.4$, $w_2=0.2$, and $w_3=0.4$, respectively.
Here, we describe the details of the proposed IPRM and RCM.
\subsection{Interclass Prototype Relation Module}
The Interclass Prototype Relation Module (IPRM) was proposed to reduce the similarity between classes. The IPRM has a prototype generation process and relation loss calculation process.
The prototype generation process calculates the prototypes of all classes present in the batch and the support images.
This process obtains a prototype, which is a global descriptor of a particular class in an image by taking as input the feature map extracted from the support image and the segmentation mask paired with it.
We employed the masked average pooling strategy~\cite{AMP} to compute the prototype for each class. The prototype $P_i^c$ of the $c$ th class in the $i$ th support image is computed using equation (\ref{eq:prototype}).
\begin{align}
\label{eq:prototype}
P_i^c=\frac{\sum_{x,y}F_{i}^{x,y}\mathbbm{1}[M_{i}^{x,y}=c]}{\sum_{x,y}\mathbbm{1}[M_{i}^{x,y}=c]}
\end{align}
$F_i$ is the feature map extracted via the backbone network with the support image $I_i^S$ as the input.
The subscripts $x$ and $y$ indicate the horizontal and vertical spatial positions of the feature map, respectively.
$M_{i}^{x,y}$ is the segmentation mask. By applying this, we can eliminate the regions of the feature map other than the specified class.
Equation (\ref{eq:prototype}) was calculated for all the support images in the batch. The maximum number of prototypes $n$ obtained is the number of all classes in which $S_{train}$ has $c\in\{ 0, 1, 2, \ldots, n \}$.

Next, the relation loss calculation process was performed using the obtained prototypes.
This process calculates the average value $L_r$ of the cosine similarity between different classes.
The relation loss $L_r$ is calculated using equation (\ref{eq:loss},\ref{eq:sim}).
\begin{align}
\label{eq:loss}
L_r=\frac{\sum_{c_s}^n\sum_{c_t}^n Sim(P^{c_s}, P^{c_t})
\mathbbm{1}[c_{s}\neq c_{t}]}
{\sum_{c_s}^n\sum_{c_t}^n \mathbbm{1}[c_{s}\neq c_{t}]}
\end{align}

\begin{align}
\label{eq:sim}
Sim(P^{c_s}, P^{c_t}) = \frac{ {P^{c_{s}}} \cdot {P^{c_{t}} } }
{ \norm{{P^{c_{s}}}} \cdot \norm{P^{c_{t}}} }
\end{align}
The $c_s$ and $c_t$ refer to class numbers and the difference between the prototypes of two different classes is measured by the cosine similarity expressed in the equation (\ref{eq:sim}).
The similarity between the prototypes of two different classes $c_s$,$c_t$ computed in equation (\ref{eq:sim}) is calculated for all combinations switching between pairs of prototypes to be measured using the equation (\ref{eq:loss}).
The average of these prototype similarity values are the relation loss $L_r$.
The relation loss $L_r$ is designed to improve the separation performance between each class by training the network such that the similarity between each class is low.

Further, among the prototypes calculated by the equation (\ref{eq:prototype}), the prototype of the target class and the prototype extracted from the background region is selected.
The prototypes were compared with the query feature and their respective similarity maps were computed. These similarity maps were used as input to the multi-scale network.
\subsection{Respective Classifier Module}
\begin{figure}[h]
  \includegraphics[width=1.0\textwidth]{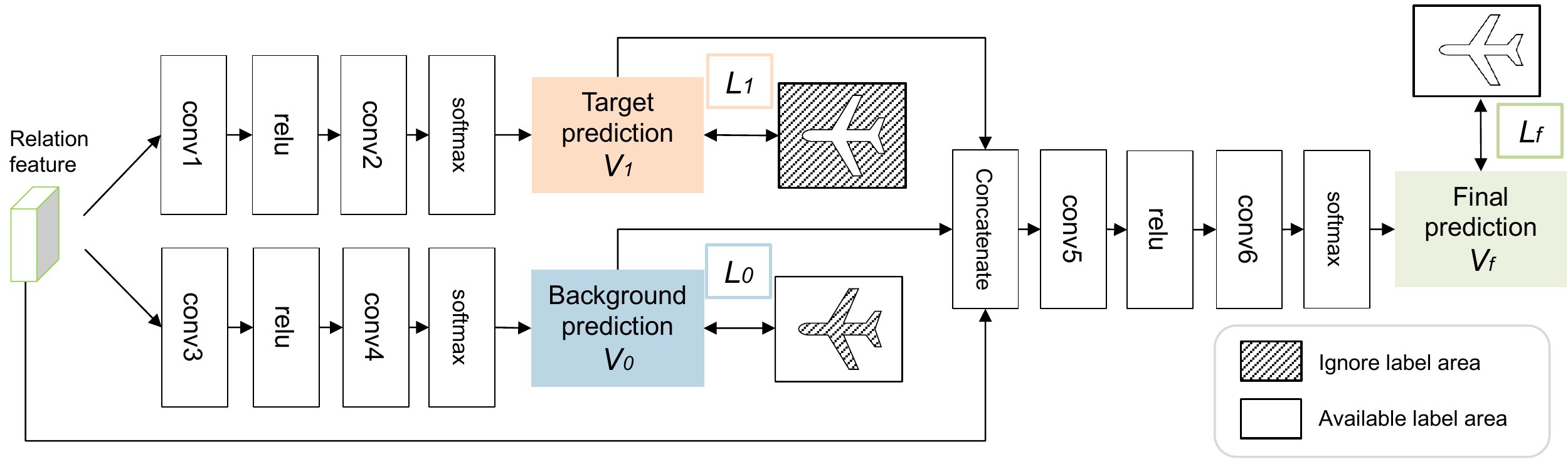}
  \caption{Configuration diagram of the Respective Classifier Module (RCM).}
  \label{fig:rcm}
\end{figure}

An overview of the Respective Classifier Module (RCM) is shown in Figure \ref{fig:rcm}.
The RCM takes as input the relation feature, which is the output of the multi-scale network shown in Figure \ref{fig:overrall}.
It was designed to improve the separation performance between the target class of objects and the background by estimating each of them independently, reintegrating the results.
The relation feature $F_r$ is branched for foreground target class prediction and background prediction. Then, it is transformed into a probability distribution representation $V_1$ and $V_0$ through two convolutional layers, the activation layer, and the softmax layer respectively.
$V_1$ indicates the probability distribution of a single object in the target class, and $V_0$ indicates the probability distribution of the background region only.
The loss calculation between the probability distribution representation $V$ and the ground-truth mask $M^Q$ can be expressed by equation (\ref{eq:l_f}).
\begin{align}
\label{eq:l_f}
L_c=\frac{-\sum_{x,y}D_c(M^Q)log(V_c^{x,y})}
{\sum_{x,y}D_c(M^Q)}
\end{align}
\begin{align}
\label{eq:mask}
D_c(M^Q)=
\begin{cases}
    c   & \text{if $M^Q=c$,} \\
    255 & \text{if $M^Q\neq c$.}
\end{cases}
\end{align}
The loss is calculated by cross-entropy error.
$V_1$, the calculated loss value $L_1$ is the loss function of a single object of the target class, and given probability distribution $V_0$, the calculated loss value $L_0$ is the loss function of the background region only.
Equation (\ref{eq:mask}) is a function that, according to the given class ID $c$, returns 255 pixel positions other than those corresponding to the class ID in the ground-truth mask $M^Q$.
Two hundred and fifty-five means ignore the label, an area that is not used when calculating loss, thus eliminating the relationship between the background and foreground.
This mechanism allows the RCM to acquire a discriminator through training that can make inferences about its pixels based on information about the object itself only.

The final estimation result $V_f$ can be obtained by fusing the information from the concatenation of the foreground probability distribution $V_1$, background probability distribution $V_0$, and relation feature $F_r$ with the two convolution layers, the activation layer, and the softmax layer.
The cross-entropy error $L_f$ was also obtained between the prediction result $V_f$ and the ground-truth mask ${M^Q}$.

The final loss value $L_p$ output by the RCM is obtained by the weighted addition of the loss value $L_1$ to the prediction result of the foreground object alone, the loss value $L_0$ to the prediction result of the background region only, and the loss value $L_f$ to the final prediction result. This can be defined by the following equation (\ref{eq:l_p}).
\begin{align}
\label{eq:l_p}
L_p = \alpha L_1 + \beta L_0 + \gamma L_f  
\end{align}
$\alpha$, $\beta$, and $\gamma$ are the weight coefficients, which were trained by setting $\alpha=0.15$, $\beta=0.15$, and $\gamma=0.7$, respectively.
\section{Experiments}
\subsection{Datasets and Evaluation Metric}
To analyze the performance of IPRNet, we selected two datasets that are widely used for few-shot segmentation, Pascal-$5^i$\cite{Oneshot} and COCO-$20^i$~\cite{coco}.
Pascal-$5^i$~\cite{Oneshot} contains images from PASCAL VOC 2012~\cite{pascal} and additional annotations from SBD~\cite{SBD}.
For training, a total of twenty class categories were evenly divided into four splits, and model training was performed using a cross-validation method.
Specifically, three splits were selected as the training data during the training process, and the remaining splits were used for testing.
During testing, one thousand support-query pairs were randomly sampled and evaluated~\cite{Oneshot}.
COCO-$20^i$~\cite{coco}, unlike Pascal-$5^i$\cite{Oneshot}, is an incredibly challenging dataset. This is because it is a large dataset having 82,081 images, and many objects are included in the images of realistic scenes.
Following the FWB~\cite{FWB}, the eighty classes of COCO-$20^i$~\cite{coco} were evenly divided into four splits, and the same cross-validation scheme was used. To obtain more stable results, we randomly sampled 20,000 pairs of during the testing~\cite{PFENet}.
As an evaluation metric, we used the mean Intersection-over-Union (mIoU), which is commonly used in semantic segmentation.

An ablation study was conducted to verify the influence of the proposed IPRM and RCM. To verify the separation performance between similar classes, we compared the per-class IoU results of IPRNet and the baseline without our IPRM and RCM.
\subsection{Implementation details}
ResNet~\cite{ResNet} is employed as the backbone network, and block2 and block3 are concatenated to generate the feature map~\cite{CANet}.
The input support and query images were cropped to an image size of 400×400 pixels and fed into the backbone network.
The initial learning rate was set to 0.05, momentum and weight decay to 0.9 and 0.0001, respectively, and the optimizer was trained with a batch size of thirty-two using the SGD optimizer.
We also adapted the poly method~\cite{Deeplab}, where the decay of the learning rate is achieved by multiplying by $(1-current\_iter )^{power}$, and the $power$ is set to 0.9.
The pretrained backbone network is frozen so that it does not learn class-specific representations of the training data.
We implemented it using Pytorch experimented with an NVIDIA A10G GPU.

In the ablation study, for the IPRM deletion, we also modified a mechanism that uses only the conventional masked average pooling strategy~\cite{AMP} to acquire the target class prototype and background class prototype.
The RCM is completely removed and replaced with a mechanism that calculates the loss between the output of the multi-scale network and the ground truth.
\subsection{Experimental results}
The effectiveness of our method was evaluated on two benchmark datasets Pascal-$5^i$~\cite{Oneshot} and COCO-$20^i$~\cite{coco}.
We extensively experimented with the $1$-shot and $5$-shot on $1$-way problems in various few-shot split settings using the widely used encoder networks ResNet-50 and ResNet-101.
Here, an n-way k-shot implies that k samples are given for each class among the n classes.
Extensive experiments show that the mIoU is improved over the conventional method in all the cases.
The ablation study confirmed that removing the IPRM and RCM reduced the mIoU.
\subsubsection{Pascal-$5^i$}
First, we describe the results for Pascal-$5^i$ in Table \ref{tb:pascal}. Our method has
an improved mIoU in all conditions compared with HSNet~\cite{HSNet}, which has the highest mIoU among the conventional methods.
In order of performance improvement, we first see a 1.7\% mIoU improvement for ResNet-50 and 1.3\% mIoU improvement for ResNet-101 in the 1-shot setting. These are followed by 0.7\% mIoU improvement for ResNet-50 and 0.5\% mIoU improvement for ResNet-101 in the 5-shot setting.
\subsubsection{COCO-$20^i$}
Next, we describe the results for COCO-$20^i$ in Table \ref{tb:coco}.
Our method has improved mIoU in all conditions compared with HSNet~\cite{HSNet}, which has the highest mIoU among the conventional methods.
In order of performance improvement, we first see a 6.1\% mIoU improvement for ResNet-50 and 5.7\% mIoU improvement for ResNet-101 in the 1-shot setting. This is followed by a 4.2\% mIoU improvement for ResNet-50 and 3.8\% mIoU improvement for ResNet-101 in the 5-shot setting.
\begin{table}[htb]
\centering
  \caption{
   Performance on Pascal-$5^i$ in IoU with per-splits results. Some results are from~\cite{PANet, PPNet, PFENet, ASGNet, MLC, HSNet}.
   }
   \label{tb:pascal}
  \scalebox{0.85}[0.85]{ 
  \begin{tabular}{l|l|ccccc|ccccc} \toprule

    \multirow{2}{*}{Backbone} & \multirow{2}{*}{Method} & \multicolumn{5}{|c|}{1shot} & \multicolumn{5}{|c}{5shot} \\  
    
     & &\, s-0 \,  &\, s-1 \, &\, s-2 \, &\, s-3 \, &\, mean \, &\, s-0 \, &\, s-1 \, &\, s-2 \, &\, s-3 \, &\, mean  \\  \midrule
    
    \multirow{7}{*}{ResNet-50} 
    
    & PANet~\cite{PANet} 
    &44.0&57.5&50.8&44.0&49.1&55.3&67.2&61.3&53.2&59.3\\
    
    & PPNet~\cite{PPNet} 
    &48.6&60.6&55.7&46.5&52.8&58.9&68.3&66.8&58.0&63.0\\
    
    & PFENet~\cite{PFENet} &61.7&69.5&55.4&56.3&60.8&63.1&70.7&55.8&57.9&61.9\\
    
    & ASGNet~\cite{ASGNet} 
    &58.8&67.9&56.8&53.7&59.3&63.7&70.6&64.2&57.4&63.9\\

    & MLC~\cite{MLC} 
    &59.2&71.2&\textbf{65.6}&52.5&62.1&63.5&71.6&\textbf{71.2}&58.1&66.1\\
    
    & HSNet~\cite{HSNet}&64.3&70.7&60.3&60.5&64.0&\textbf{70.3}&73.2&67.4&\textbf{67.1}&69.5\\
    
    & \textbf{Ours} &\textbf{65.2}&\textbf{72.9}&63.3&\textbf{61.3}&\textbf{65.7}&70.2&\textbf{75.6}&68.9&66.2&\textbf{70.2}\\

    \hline \midrule
    \multirow{7}{*}{ResNet-100} 
    
    & FWB~\cite{FWB}
    &51.3&64.5&56.7&52.2&56.2&54.8&67.4&62.2&55.3&59.9\\     
    
    & PPNet~\cite{PPNet} 
    &52.7&62.8&57.4&47.7&55.2&60.3&70.0&69.4&60.7&65.1\\
    
    & PFENet~\cite{PFENet}
    &60.5&69.4&54.4&55.9&60.1&62.8&70.4&54.9&57.6&61.4\\
    
    & ASGNet~\cite{ASGNet}
    &59.8&67.4&55.6&54.4&59.3&64.6&71.3&64.2&57.3&64.4\\

    & MLC~\cite{MLC} &60.8&71.3&61.5&56.9&62.6&65.8&74.9&71.4&63.1&68.8\\

    & HSNet~\cite{HSNet}
    &67.3&72.3&62.0&\textbf{63.1}&66.2&\textbf{71.8}&74.4&67.0&\textbf{68.3}&70.4\\

    & \textbf{Ours} &\textbf{67.8}&\textbf{74.6}&\textbf{65.7}&62.2&\textbf{67.5}&70.0&\textbf{75.9}&\textbf{71.8}&65.8&\textbf{70.9}\\

    \bottomrule
    
  \end{tabular}
  }
\end{table}
\begin{table}[htb]
\centering
  \caption{
   Performance on COCO-$20^i$ in IoU with per-splits results. Some results are from~\cite{PPNet, PFENet, MLC, HSNet}. 
   }
   \label{tb:coco}
  \scalebox{0.85}[0.85]{ 
  \begin{tabular}{l|l|ccccc|ccccc} \toprule

    \multirow{2}{*}{Backbone} & \multirow{2}{*}{Method} & \multicolumn{5}{|c|}{1shot} & \multicolumn{5}{|c}{5shot} \\  
    
     & &\, s-0 \,  &\, s-1 \, &\, s-2 \, &\, s-3 \, &\, mean \, &\, s-0 \, &\, s-1 \, &\, s-2 \, &\, s-3 \, &\, mean  \\  \midrule
    
    \multirow{5}{*}{ResNet-50}
    & PPNet~\cite{PPNet} 
    &28.1&30.8&29.5&27.7&29.0&39.0&40.8&37.1&37.3&38.5 \\
    
    & PFENet~\cite{PFENet} 
    &36.5&38.6&34.5&33.8&35.8&36.5&43.3&37.8&38.4&39.0 \\
    
    & MLC~\cite{MLC}    
    &\textbf{48.0}&36.6&27.4&28.2&35.1&\textbf{54.9}&42.1&34.9&33.6&41.4 \\
    
    & HSNet~\cite{HSNet}    
    &36.3&43.1&38.7&39.2&39.2&43.3&51.3&48.2&45.0&46.9 \\

    & \textbf{Ours} &42.2&\textbf{48.9}&\textbf{45.5}&\textbf{44.6}&\textbf{45.3}&48.0&\textbf{55.7}&\textbf{50.7}&\textbf{50.1}&\textbf{51.1}\\    
    
    \hline \midrule
    \multirow{5}{*}{ResNet-100} 
    &FWB~\cite{FWB}
    &17.0&18.0&21.0&28.9&21.2&19.1&21.5&23.9&30.1&23.7 \\
    
    & PFENet~\cite{PFENet} 
    &34.3&33.0&32.3&30.1&32.4&38.5&38.6&38.2&34.3&27.4 \\
    
    & MLC~\cite{MLC}    
    &\textbf{51.1}&38.7&28.5&31.6&37.5&\textbf{57.8}&47.1&37.8&37.6&45.1 \\
    
    & HSNet~\cite{HSNet}  
    &37.2&44.1&42.4&41.3&41.2&45.9&53.0&51.8&47.1&49.5 \\
    
    & \textbf{Ours} &42.9&\textbf{50.6}&\textbf{46.8}&\textbf{47.4}&\textbf{46.9}&50.7&\textbf{58.3}&\textbf{52.8}&\textbf{51.3}&\textbf{53.3}\\

    \bottomrule
    
  \end{tabular}
  }
\end{table}
\subsubsection{Ablation study}
We conducted an ablation study to investigate the influence of the IPRM and RCM, which are the main components of our model.
All ablation study experiments were the result of a 5-shot setup performed on the COCO-$20^i$ dataset using the ResNet50 backbone. The results are presented in Table \ref{tb:ab}.
The largest decrease in mIoU is seen in the case where both IPRM and RCM are removed by 4.4\%, followed by the case where only IPRM is removed by 2.5\%, and finally, the case where only RCM is removed 1.8\%. The mIoU decreased by more than 0.7\% when the IPRM was removed then when the RCM was removed.
\begin{table}[htb]
\centering
  \caption{Influence of the IPRM and RCM on COCO-$20^i$ in IoU with 5-shot ResNet-50 backbone condition.}
  \label{tb:ab}
  \scalebox{0.92}[0.92]{ 
  \begin{tabular}{ccccccc} \toprule
     IPRM \, &\, RCM \, &\, s-0 \, &\, s-1 \, &\, s-2 \, &\, s-3 \, &\, Mean \, 
     \\  \midrule
                    & & 42.0 & 52.1 & 46.4 & 46.4 & 46.7 \\
     \CheckmarkBold &  & 46.8 & 53.6 & 47.9 & 49.0 & 49.3 \\ 
     &\CheckmarkBold& 44.9 & 53.2 & 47.4 & 48.9 & 48.6 \\

     \midrule  
     
     \CheckmarkBold & \CheckmarkBold & \textbf{48.0} & \textbf{55.7} & \textbf{50.7} & \textbf{50.1} & \textbf{51.1} \\ \bottomrule
  \end{tabular}}
\end{table}
\subsubsection{IOU of each class}
To verify whether our proposed method is effective for objects with similar features that are difficult to classify, we evaluated it using COCO-$20^i$ is an incredibly challenging dataset that has many objects in the images of realistic scenes. 
The evaluation is perfomed by comparing IoUs per class with the baseline, which eliminates the IPRM and RCM, and our proposed IRPNet.
The results are presented in Table \ref{tb:iou}.
Significant improvements in IoU were as follows 1 person 20.7\%, 61 dining table 20.2\%, 73 fridge 15.4\%, 66 remote 13.4\%, 67 keyboard 12.3\%, 27 handbag 11.6\%, 29 suitcase 11.5\%, 25 backpack 11.3\%, 74 book 11.3\%, 44 knife 10.3\%; and an increase of more than 10\% in IoU.
The classes with a lower IoU were 34 kite -8.9\%, 65 mouse -8.2\%, 48 apple -4.6\%, 59 potted plant -3.8\%, 13 park meter -3.5\%, 4 motorcycle -2.8\%, 10 traffic light -1.8\%, 26 umbrella -1.5\%, 69 microwave -1.5\%, 80 toothbrush -1.0\%; and a decrease of over 1\% in IoU.
The following is a discussion.
\begin{table}[htb]
\centering 
\caption{COCO-$20^i$ performance in all classes of IoU experimented with 5shot ResNet-50. Baseline means the result of eliminating the IPRM and RCM.}
\label{tb:iou}
  \scalebox{0.63}[0.63]{ 
  \begin{tabular}{lcc|lcc|lcc|lcc} \hline
     \multicolumn{1}{c}{s-0} &\, Baseline \, & \, \textbf{Ours} \,
    &\multicolumn{1}{c}{s-1} &\, Baseline \, & \, \textbf{Ours} \,
    &\multicolumn{1}{c}{s-2} &\, Baseline \, & \, \textbf{Ours} \,
    &\multicolumn{1}{c}{s-3} &\, Baseline \, & \, \textbf{Ours} 
    \\ \hline \hline
1 Person& 30.3 & \textbf{51.0} &2 Bicycle& 52.9 & \textbf{55.8} &3 Car& 35.7 & \textbf{38.5} &4 Motorcycle& \textbf{56.0} & 53.2 \\
5 Airplane& 73.0 & \textbf{76.0} &6 Bus& 69.0 & \textbf{72.2} &7 Train& 72.1 & \textbf{72.8} &8 Truck& \textbf{35.1} & 34.7 \\
9 Boat& 40.9 & \textbf{50.5} &10 T.light& \textbf{40.8} & 39.0 &11 Fire H.& 77.4 & \textbf{82.7} &12 Stop sign& 76.5 & \textbf{81.4} \\
13 Park meter& \textbf{60.1} & 56.6 &14 Bench& 35.7 & \textbf{38.3} &15 Bird& 64.5 & \textbf{69.0} &16 Cat& 77.4 & \textbf{82.0} \\
17 Dog& 65.0 & \textbf{73.6} &18 Horse& 70.6 & \textbf{74.7} &19 Sheep& 75.2 & \textbf{76.8} &20 Cow& 73.0 & \textbf{78.7} \\
21 Elephant& 79.7 & \textbf{83.0} &22 Bear& 83.6 & \textbf{85.6} &23 Zebra& 75.9 & \textbf{76.2} &24 Giraffe& 72.6 & \textbf{75.6} \\
25 Backpack& 18.5 & \textbf{29.8} &26 Umbrella& \textbf{60.0} & 58.5 &27 Handbag& 21.2 & \textbf{32.8} &28 Tie& 17.8 & \textbf{18.6} \\
29 Suitcase& 42.7 & \textbf{54.2} &30 Frisbee& 69.6 & \textbf{75.9} &31 Skis& 31.3 & \textbf{38.7} &32 Snowboard& 37.4 & \textbf{46.3} \\
33 Sports ball& 41.3 & \textbf{48.4} &34 Kite& \textbf{51.3} & 42.4 &35 B. bat& 31.0 & \textbf{35.1} &36 B. glove& 48.3 & \textbf{50.4} \\
37 Skateboard& \textbf{42.4} & \textbf{42.4} &38 Surfboard& 64.7 & \textbf{68.8} &39 T.racket& 58.4 & \textbf{65.3} &40 Bottle& 28.4 & \textbf{32.6} \\
41 W. glass& 34.0 & \textbf{37.6} &42 Cup& 49.8 & \textbf{56.6} &43 Fork& 19.7 & \textbf{22.7} &44 Knife& 34.8 & \textbf{45.1 }\\
45 Spoon& 14.2 & \textbf{16.8} &46 Bowl& 31.1 & \textbf{31.9} &47 Banana& 41.6 & \textbf{45.5} &48 Apple& \textbf{39.4} & 34.8 \\
49 Sandwich& 50.5 & \textbf{52.8} &50 Orange& 47.5 & \textbf{49.5} &51 Broccoli& 33.1 & \textbf{36.8} &52 Carrot& 23.1 & \textbf{27.3} \\
53 Hot dog& 61.7 & \textbf{67.5} &54 Pizza& 84.7 & \textbf{87.5} &55 Donut& 67.6 & \textbf{70.8} &56 Cake& 44.1 & \textbf{51.9} \\
57 Chair& 7.2 & \textbf{14.2} &58 Couch& 34.5 & \textbf{37.0} &59 P. plant& \textbf{11.9} & 8.1 &60 Bed& 53.7 & \textbf{56.8} \\
61 D.table& 27.5 & \textbf{47.7} &62 Toilet& 57.2 & \textbf{63.6} &63 TV& 44.5 & \textbf{52.2} &64 Laptop& 50.9 & \textbf{55.4} \\
65 Mouse& \textbf{50.1} & 41.9 &66 Remote& 45.4 & \textbf{58.8} &67 Keyboard& 21.0 & \textbf{33.3} &68 Cellphone& 56.2 & \textbf{65.8} \\
69 Microwave& \textbf{34.9} & 33.4 &70 Oven& 23.6 & \textbf{29.7} &71 Toaster& \textbf{42.4} & 41.8 &72 Sink& 30.8 & \textbf{33.0} \\
73 Fridge& 23.9 & \textbf{39.3} &74 Book& 10.2 & \textbf{21.5} &75 Clock& 59.6 & \textbf{64.6} &76 Vase& 41.7 & \textbf{46.1} \\
77 Scissors& 43.4 & \textbf{43.8} &78 Teddy& 60.6 & \textbf{64.2} &79 Hairdrier& 47.9 & \textbf{49.4} &80 Toothbrush& \textbf{34.3} & 33.3 
\\ \hline
  \end{tabular}
  }
\end{table}
\subsection{Discussion}
\subsubsection{Comparison with State-of-the-Arts}
We discuss the mIoU of our IPRNet and the latest and best performing HSNet~\cite{HSNet}.
By comparing the results of Pascal-$5^i$ and COCO-$20^i$ described in Table \ref{tb:pascal} and \ref{tb:coco}, we can see that the performance of the COCO-$20^i$ condition is better than that of the Pascal-$5^i$ condition, regardless of the backbone network and number of shots.
We discuss that our mechanism of training to avoid similarity between prototypes is effective for this difficult problem; COCO-$20^i$, contains many more difficult objects to classify.
Comparing the experimental results of 1-shot and 5-shot, the mIoU of 1-shot is higher regardless of the backbone network and dataset.
This is because, as mentioned in the introduction, when the number of shots is smaller, the feature space covered by the support data is sparser and the classification performance was worse for classes near the classification boundary. Therefore, we can conclude that the proposed method is more effective.
\subsubsection{Performance of similar classes that are difficult to classify}
For IPRNet and the baseline without the IPRM and RCM, we compare and discuss IoU for each of the classes shown in Table \ref{tb:iou}.
Two main cases of classes that are difficult to classify are considered because of the existence of similar classes.

\begin{figure}[htb]
  \includegraphics[width=1.0\textwidth]{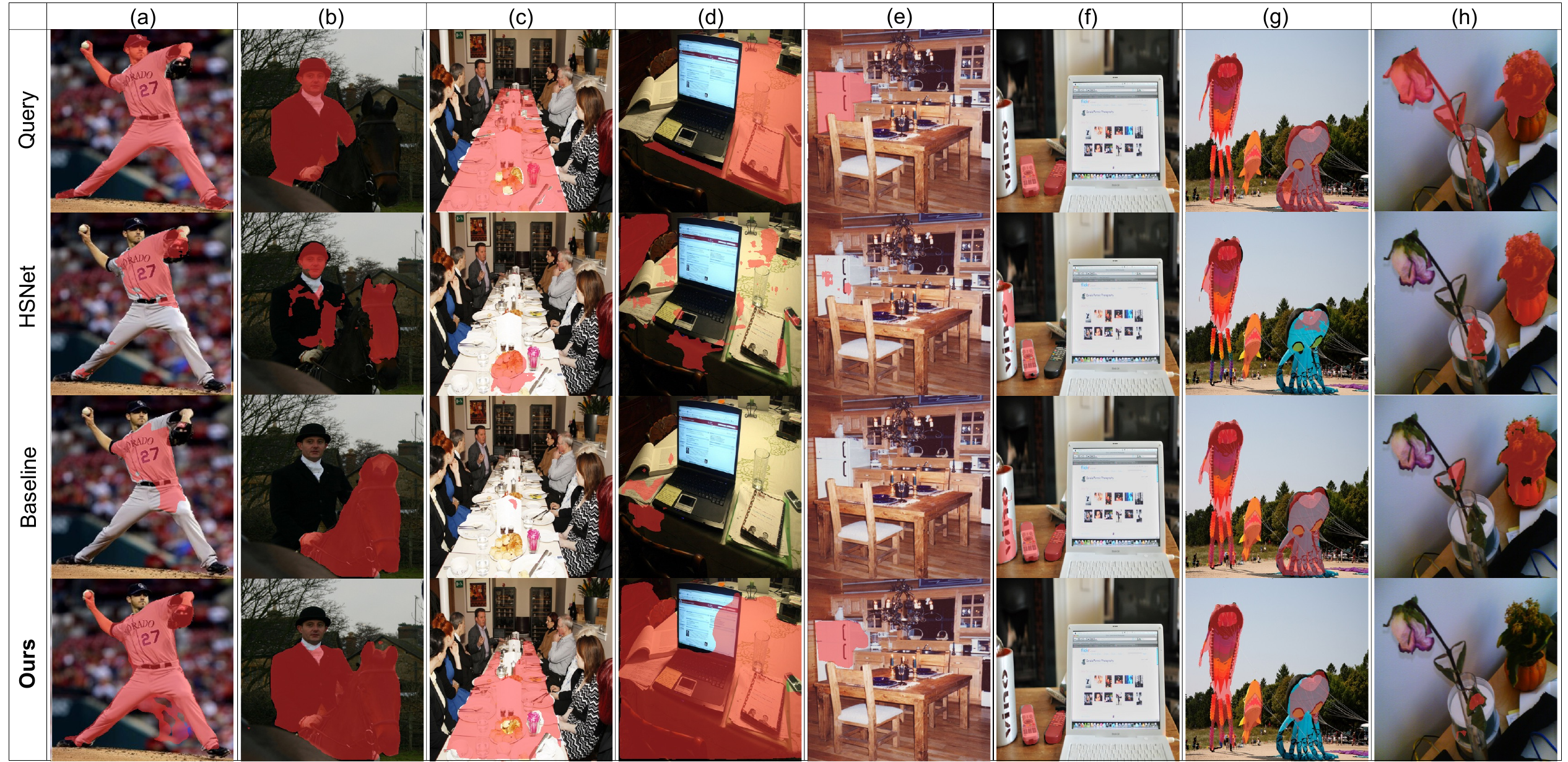}
  \caption{Examples of the recognition result using our proposed method, baseline and HSNet~\cite{HSNet}. The red shade is the area of the target class. It is the ground truth in the case of query, the recognition result in the cases of the baseline, HSNet~\cite{HSNet} and Ours.}
  \label{fig:com_img}
\end{figure}

The first is an object that is likely to be similar to many other classes with a wide range of variations, specifically morphological changes, pictorial changes through decoration, and combinations with complex backgrounds.
Specifically, there are 1 person with IoU increased by 20.7\% and 61 dining table with IoU increased by 20.2\%.
1 Person has all types of shape changes due to a deformable body and pictorial changes due to decoration (Fig.\ref{fig:com_img}(a),(b)).
61 Dining table is a complex combination of objects, most of which are placed on top of each other; therefore, the pixel boundaries of the objects are always shared with various occlusions (Fig.\ref{fig:com_img}(c),(d)).

Second, the class is an object with a simple shape and few features.
Specifically, 73 fridge IoU increased by 15.4\%, 66 remote by 13.4\%, 67 keyboard by 12.3\%, 27 handbag by 11.6\%, 29 suitcase by 11.5\%, 25 backpack by 11.3\%, 74 book by 11.3\%, and 44 knife by 10.3\%.
For example, 73 fridge is a symmetrical rectangular object with a few prominent patterns or protrusions on its surface. Hence, it was difficult to classify them based on similar rectangles and backgrounds (Fig.\ref{fig:com_img}(e)).
66 Remote is also a small rectangular body (Fig.\ref{fig:com_img}(f)). 

For all other classes with an improved IoU, there are many classes where the IoU has increased because the RCM effect is considered to have improved the separation performance from the background.
However, compared to the classes where these performances have increased by over 10\%, the degree of conformity between the two cases are considered to be low.
Specifically, 45 spoon and 43 fork are considered not to have improved IoU by as much as 44 knife because the shape of the tip is more non-graphical and distinctive compared to 44 knife.
\subsubsection{Performance of characteristic objects}
Further, we consider the classes with the lowered IoU as proper nouns that have a complex shape or structure unique to that object that is unparalleled.
Specifically, there are 80 toothbrush, 69 microwave, 26 umbrella, 10 traffic light, 4 motorcycle, 13 park meter, 59 potted plant, 48 apple, 65 mouse and 34 kite.
For example, 34 kite is a discriminative proper nouns with no other similar concepts and limited use, so it is not often combined with several backgrounds (Fig.\ref{fig:com_img}(g)).
9 Potted plants had complex shapes that could not be represented by rectangles or spheres (Fig.\ref{fig:com_img}(h)).
For these objects, we consider that how to extract the unique features of the object to be more important than acquiring the differences from other classes, and the training to reduce the similarity of prototypes between different classes, which is the aim of the IPRM, does not work effectively.
\subsubsection{Qualitative Evaluation}
\begin{figure}[h]
  \includegraphics[width=1.0\textwidth]{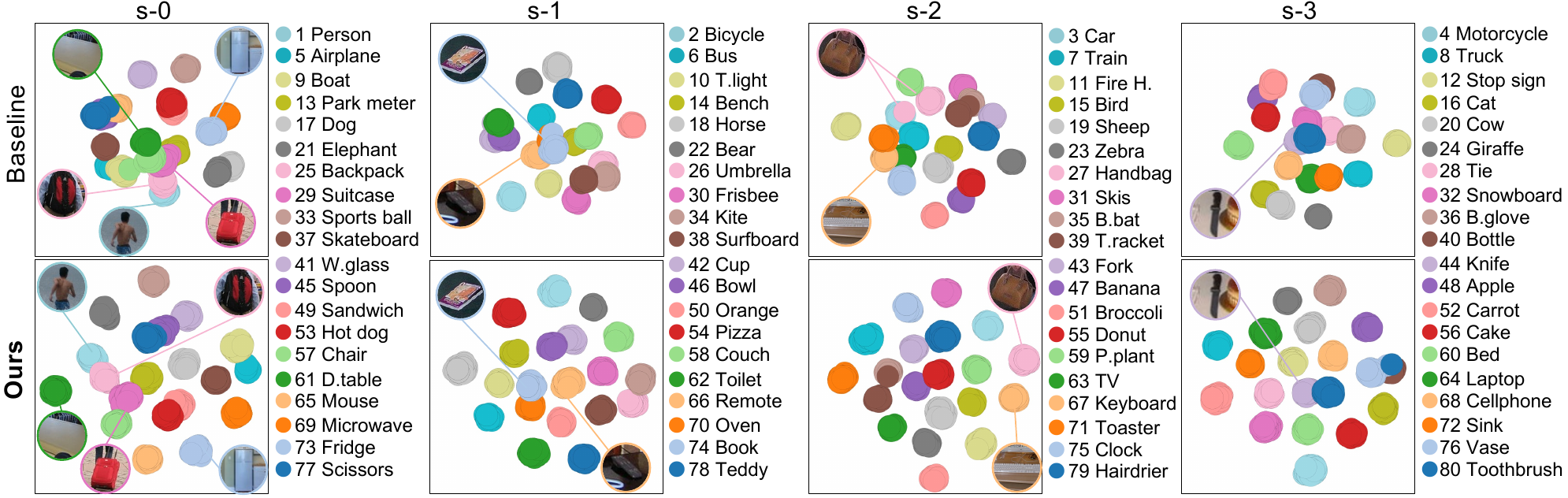}
  \caption{Visualization results by t-SNE~\cite{tsne} of prototype obtained from query image of the baseline and IPRNet.
  They experiment with ResNet-50 5shot setting on COCO-$20^i$.}
  \label{fig:tsne}
\end{figure}

Fig.\ref{fig:tsne} shows the difference between the baseline and our IPRNet using t-SNE~\cite{tsne} for the prototype of the target class extracted from the query image.
At the baseline, prototypes between different classes are adjacent or overlapping, and there is no clear classification. 
However, in IPRNet, the distance between each prototype is increased, and it can be observed that the prototypes are more separated.
\section{Conclusion}
This study proposes a novel IPRNet that introduces a mechanism to improve separation performance by reducing the similarity between different classes.
Experiments show that mIoU is improved over the existing few-shot segmentation methods~\cite{PPNet}~\cite {PFENet}~\cite {MLC}~\cite{HSNet}. 
In addition, through an ablation study, we verified the separation performance between similar classes that are difficult to classify.
\paragraph{Acknowledgments}
We would like to thank Mr.Katsushi Yamashita, Director, from R\&D Promotion Office, SoftBank Corp. and  all of R\&D Promotion Office members. Mr.Yamashita's witty and helpful support helped us, especially.
We would like to thank Mr.Shogo Hamano from Department of Mechano-Informatics, the University of Tokyo for his informative discussion.
\clearpage
%
%
\bibliographystyle{splncs04}
\bibliography{egbib}
\end{document}